\documentclass[10pt,twocolumn,letterpaper]{article}
\pdfoutput=1
\usepackage{iccv}
\usepackage{times}
\usepackage{epsfig}
\usepackage[noadjust]{cite}

\usepackage[usenames,dvipsnames,svgnames,table]{xcolor}
\usepackage{graphicx}
\usepackage{amsmath,amssymb}
\usepackage{amsthm}
\usepackage{bm}
\usepackage{tabularx}
\usepackage{multirow}
\usepackage{hhline}
\usepackage{subfig}
\usepackage{booktabs}
\usepackage{sidecap}
      
\usepackage[normalem]{ulem}

\usepackage[pagebackref=true,breaklinks=true,letterpaper=true,colorlinks,bookmarks=false]{hyperref}

\def\ie{\emph{i.e.}\xspace}

\newcommand{\mb}[1]{\bm #1}

\newtheorem{theorem}{Theorem}
\newtheorem{proposition}[theorem]{Proposition}
\newcommand{\eqr}[1]{Eq. \eqref{#1}}
\newcommand{\figurecorrection}{\vspace{-5mm}}

\iccvfinalcopy 

\begin{document}

\title{\emph{Active Transfer Learning with Zero-Shot Priors}:\\ Reusing Past Datasets for Future Tasks}

\author{E. Gavves\thanks{This work has been done while working at ESAT-PSI, KU Leuven.}\\
QUVA Lab, University of Amsterdam\\
%
\and
T. Mensink\\
University of Amsterdam\\
%
\and
T. Tommasi\footnotemark[1]\\
UNC Chapel Hill\\
%
\and
C.G.M. Snoek\\
QUVA Lab, University of Amsterdam\\
%
\and
T. Tuytelaars\\
KU Leuven, ESAT-PSI, iMinds\\
}

\maketitle

\begin{abstract}
How can we reuse existing knowledge, in the form of available  datasets, when solving a new and apparently unrelated target task from a set of unlabeled data? 
In this work we make a first contribution to answer this question in the context of image classification. 
We frame this quest as an active learning problem and use zero-shot 
classifiers to guide the learning process by linking the new task to the
existing classifiers.
By revisiting the dual formulation of adaptive SVM, we reveal two basic conditions to choose greedily only the most relevant samples to be annotated. 
On this basis we propose an effective active learning algorithm which learns the best possible target classification model with minimum human labeling effort. 
Extensive experiments on two challenging datasets show the value of our approach compared to the state-of-the-art active learning methodologies, as well as its potential to reuse past datasets with minimal effort for future tasks.
\end{abstract}

\section{Introduction}

\thispagestyle{empty}

Modern visual learning algorithms are founded on data. Given a set of images annotated with the desired object categories, the algorithm learns models able to recognize, detect or describe unseen images. Despite their importance, image datasets are assumed to be single-use only: whenever an object of interest is not previously annotated we urge for a new collection containing the precious label. Isn't this a wasteful approach? Take \emph{ImageNet} \cite{imagenet_cvpr09} for example. With about $15M$ images this huge collection contains much more information than the officially listed $22K$ synsets. 
However, most often such existing knowledge resources are ignored, because of no label overlap with the future tasks at hand, be it for classification~\cite{Freytag14, KovashkaVG11, Tommasi_BMVC_2012} or localization~\cite{VijayanarasimhanJG10, VezhnevetsBF12, LiECCV2014}.

\begin{figure}[t!]
    \centering
    \includegraphics[width=.9\columnwidth]{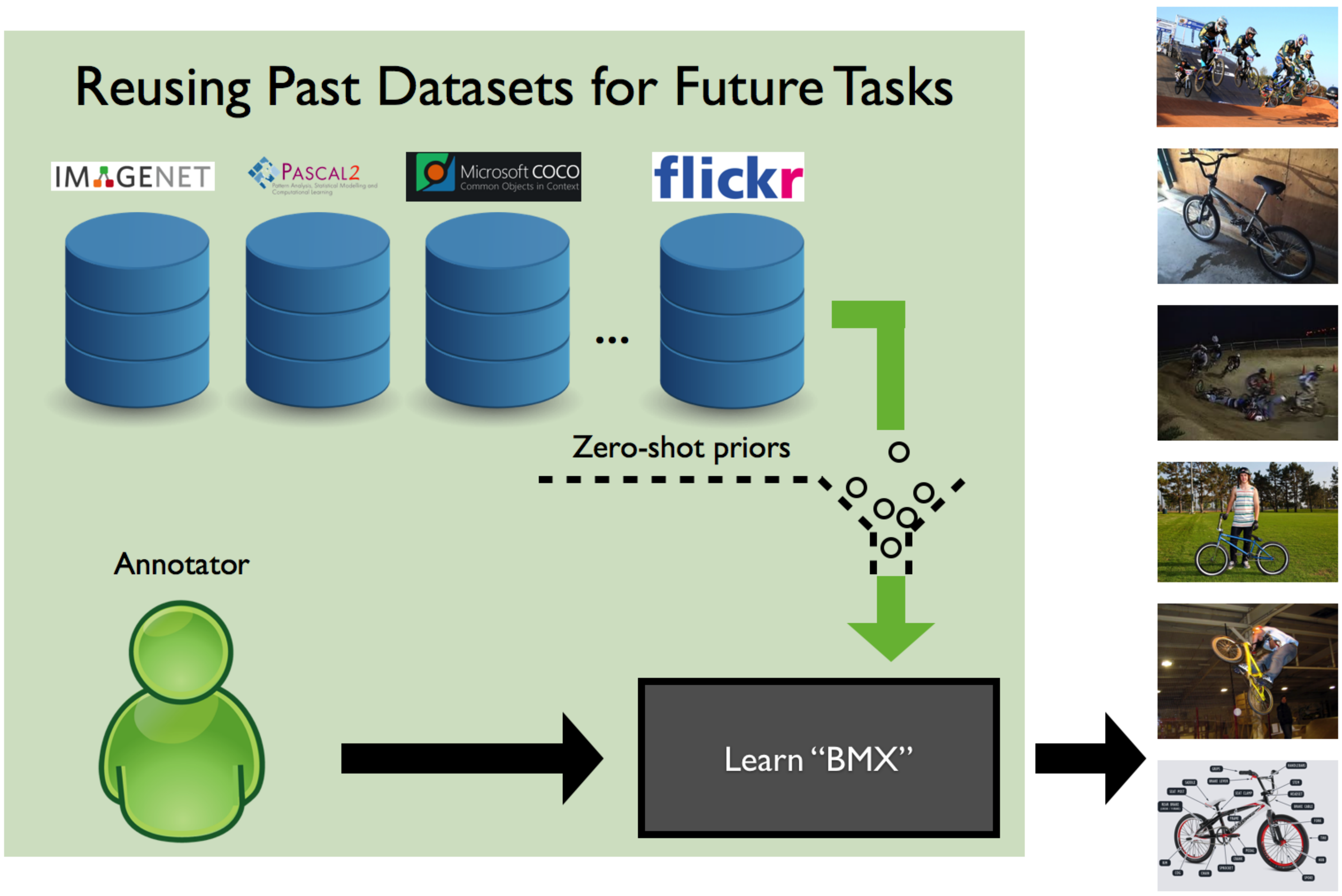}
    \caption{Imagine building a new classifier for \emph{``BMX''} bicycles, a category absent from \emph{ImageNet}. Rather than annotating the category from scratch, we propose to leverage already available annotations for learning the \emph{``BMX''} class. Through a theoretical analysis of active learning within a max-margin framework, we present the optimal conditions for sampling new data to label using zero-shot priors. \vspace{-5mm}}
\label{fig:intro}
\end{figure}

We postulate that \emph{no past knowledge is useless for future learning}, even if it appears to be so.
The fundamental question is how to reuse existing labels from given datasets, without the need for new annotations.
Previous transfer learning methods proved to be helpful when at least few annotated samples of the new target task are available~\cite{Pan_survey}. However, they were never challenged with the more difficult active learning setting, where no initial labeled data is available and the existing knowledge from the source datasets is most probably irrelevant for the new task.

We are inspired by advances in zero-shot classification~\cite{MensinkCVPR2014, lampert13pami, akata13cvpr}. Zero-shot learning was originally proposed~\cite{lampert13pami,socher_NIPS2013} as a strategy to obtain classifiers for arbitrary, novel categories, when no annotated data is available. Differently from~\cite{lampert13pami, akata13cvpr} who depend on human-provided attribute annotations, recent works demonstrate that label co-occurrences from image datasets~\cite{MensinkCVPR2014} and textual embeddings~\cite{devise, conse} can be used for reliable zero-shot classification models. Here \emph{we re-purpose zero-shot learning as priors} for a faster, more accurate and more economic active learning of novel categories. The zero-shot models provide us with some notion on the label distribution over the new unlabeled data in the feature space. By using this knowledge, and focusing on the dual formulation of SVM, we present and discuss the optimal conditions for active query sampling which we coin as the \emph{maximum conflict} and \emph{label equality} conditions. Based on these conditions we present a practical algorithm for optimal active learning sampling.


We make three contributions. First, we revisit the concept of leveraging knowledge from past datasets and generalize it: future learning benefits even from semantically unrelated existing datasets. We believe this is an interesting task both for its research potential, as well as for its relevance for practical purposes and applications. Second, we re-purpose zero-shot classifiers as zero-shot priors to guide the learning. Third, by combining zero-shot and active learning in an SVM-based framework we derive the two conditions for optimal query selection in an active learning setting. The proposed methodology is supported by an extensive evaluation on two recent datasets. Before detailing our active learning approach we first discuss related work.

\section{Related Work}
Optimizing the efficiency of artificial learning is one of the long standing goals of
computer vision. Several research directions have been proposed towards this target, with 
\emph{active}, \emph{transfer} and \emph{zero-shot learning} among the most studied and adopted 
strategies.

\textbf{Active learning.}
The objective of \emph{active learning} is to optimize a model within a limited time and annotation budget by selecting the most informative training instances~\cite{settles2009active}. A core assumption in active learning is that there is at least one positive and one negative sample for the novel category at time $t=0$.

In the context of image classification a popular pool-based active learning paradigm consists of
estimating the expected model change, and querying the samples which will likely influence the model most. 
Sampling the data closest to the support vector hyperplane~\cite{Schohn, Vijayanarasimhan2011} has 
been shown to reduce the version space of the learnt classifier~\cite{Tong2002}. 
In~\cite{VijayanarasimhanJG10} the unlabeled samples are mirrored to obtain ghost positive and negative labels.
%
%
In~\cite{Freytag14, VezhnevetsBF12} the expected model or accuracy changes are exploited to build Gaussian process classifiers.
Similarly, in~\cite{KovashkaVG11} the total entropy on the predictions is measured over all the labeled and unlabeled data, and the samples leading to the maximum entropy reduction are selected.
Other active learning methods exploit the cluster structure in the data, either by imposing a hierarchy \cite{Dasgupta} or a neighborhood graph on the unlabeled data \cite{Zhu03combiningactive} before locally propagating the labels.


\textbf{Transfer learning.} \emph{Transfer learning} aims at boosting the learning process of a new target task by transferring knowledge from previous and related source task experiences \cite{Pan_survey}. Different knowledge sources have been considered in the literature: instances \cite{lst_nips11}, models \cite{tommasi_PAMI14,Aytar12} and features \cite{Luo_ICCV11,daume}.
In the first case one exploits the availability of extra source data to enrich a poorly populated target training set.
Leveraging over models instead, allows to initialize the target learning process without the need to store the source data.
Finally, feature transfer learning relies on the source knowledge to define a representation that simplifies the target task.

Some transfer learning solutions have been proposed in a dynamic setting for online learning \cite{Tommasi_BMVC_2012}
and iterative self-labeling \cite{dasvm}. The first demonstrates the advantage of using related source knowledge
as a warm start for online target learning. The second proposes to actively annotate the most uncertain target samples with the label predicted by a known source model.

\textbf{Zero-shot learning.} Both active and transfer learning approaches suppose either the availability of at least a few labeled target training samples or an overlap with existing labels from other datasets. In contrast, \emph{zero-shot learning} exploits known relations between source and target tasks to define a target learning model without expecting annotated samples~\cite{lampert13pami,socher_NIPS2013}. This problem setting has recently attracted a lot of attention~\cite{lampert13pami, akata13cvpr, MensinkCVPR2014, devise, conse, LiECCV2014}.
Binary attributes are commonly used to encode presence or absence of visual characteristics for the object categories~\cite{lampert13pami, akata13cvpr}. They provide a description for each class and work as a natural interface for zero-shot learning. However, attribute-to-class relations are usually defined by human labelers.
Less expensive class-to-class relations can also be mined from external sources, even from textual data only~\cite{devise,conse,rohrbach10cvpr}. To avoid any labeling effort~\cite{MensinkCVPR2014} extract tag statistics from external sources like Flickr to use as class-to-class relations, whereas~\cite{Chen_2013_ICCV} directly learn novel categories by querying search engines.



All the aforementioned paradigms are closely related. With the aim of minimizing time and annotation cost, the learning process should actively select, label and use only the most relevant new data from an unlabeled set, while exploiting pre-existing knowledge. We base our approach on a max-margin formulation, which transfers knowledge from zero-shot learning models used as priors. We also present the conditions for active learning and optimal query sampling in this challenging setting.

\section{Auxiliary zero-shot active learning} \label{sub:adaptivesvm}
Let us consider a pool of unlabeled samples $\{\mb{x}_i\}_{i=0}^N \in \mathbb{R}^d$  belonging to $\mathcal{C}$ classes.
We would like to learn a classification model for each class within a limited time and annotation
budget by querying an oracle only for the labels of the most informative instances.
We consider learning a binary classifier for one of the classes with $y_i\in\{-1,+1\}$, which are not known until queried from the active learner.
We focus on standard linear classification models with the learning parameters coded into the vector $\mb{w} \in \mathbb{R}^d$.
The prediction score for class $c$ will be expressed by $f(\mb{x}) = \mb{w}\cdot\mb{x}~$ and the final annotation is obtained by $y = sign(f(\mb{x}))$.

\subsection{Maximum Conflict - Label Equality}

We can formulate a greedy learning algorithm starting from standard SVM and adding a binary selection 
variable $\gamma_i^t\in \{0, 1\}$ which indicates whether at time 
step $t$ the label $y_i$ has been queried, and therefore it is known to the classifier.
The dual objective function at time $t$ is
\begin{align}
\max_{\alpha^t, \gamma^t}&\sum_i \gamma_i^t\lambda_i^t\alpha_i^t 
                        - \frac{1}{2} \sum_{i, j} \alpha_i^t \alpha_j^t \gamma_i^t \gamma_j^t y_i y_j\mb{x}_i\cdot \mb{x}_j \label{eq:sustlearn}\\
\textrm{s.t.} \quad & \sum \gamma_i^t \alpha^t_i y_i  = 0 \label{eq:sustlearn_equilibrium}\\
                    & 0 \leq \alpha^t_i \leq C, \; \forall i ~,\label{eq:sustlearn_positivealpha}\\
                    &\gamma_i^t \geq \gamma_i^{t-1}, \; \forall i ~,\label{eq:gammacond1}\\
		    &\sum_i \gamma_i^t = \sum_i \gamma_i^{t-1} +B~ \label{eq:gammacond2}~.
\end{align}
The last two constraints define the annotation procedure over time.
\eqr{eq:gammacond1} indicates that once a sample is selected, it enters and remains in the training 
set for all the subsequent iterations. \eqr{eq:gammacond2} specifies that the number of samples 
increases in time with a budget $B$, which is the maximum annotation budget per iteration. 
According to the proposed formulation, we can define two main conditions for the optimal active
query sampling procedure. 

\begin{proposition}[\textbf{Maximum Conflict}]
To maximize the objective \eqr{eq:sustlearn} at time $t$, we should query the sample $i^*$ such that \emph{(a)} 
its label $y_{i^*}$ has an opposite sign from its classification score at $(t-1)$, 
while \emph{(b))} the classifier score is as high as possible. 
\label{prop1}
\end{proposition}
This condition follows from the role of the Lagrange multipliers $\mathbf{\alpha}^t$ in choosing the support vectors.
Intuitively, if the current model misclassifies $\mb{x}_i$, the output of the new 
classifier at time $t$ must deviate from the previous one such that it can predict $y_i$ correctly. 
This is realized by introducing in the model a new support vector $\mb{x}_i$ with a large weight $\alpha_i$.
Alternatively, if the current classifier correctly labels $\mb{x}_i$, the value of $(\mb{w}^{t}\cdot\mb{x}_i)$ 
does not need to differ from $\left(\mb{w}^{t-1}\cdot\mb{x}_i\right)$ so the weight $\alpha_i$ 
can be small or even zero, and it is less likely that $\mb{x}_i$ will be introduced in the support vector set.

\begin{proposition}[\textbf{Label Equality}]
To respect the constraint \eqr{eq:sustlearn_equilibrium} the number of positive and negative 
examples in the training set should be balanced, 
\ie $\sum_i \gamma_{i}^t  [y_i=1] = \sum_i \gamma_{i}^t  [y_i=-1]$.
\label{prop2}
\end{proposition}

By focusing only on the selected samples, from \eqr{eq:sustlearn_equilibrium} we can write:
\begin{equation}
\Big(\sum_{\forall i: \gamma^t_i=1} \alpha_i^t y_i \Big)^2=0 \Rightarrow \sum_i (\alpha_i^t)^2 + \sum_{i} \sum_{j \neq i} \alpha_i^t \alpha_j^t y_i y_j = 0.
\label{eq:equilibrium}
\end{equation}
Since for any non degenerate classifier (\ie $\mb{w} \neq \vec{0}$), holds that $\sum_i \alpha_i^2 > 0$, this implies that
\begin{equation}
\sum_{i} \sum_{j \neq i} \alpha_i \alpha_j y_i y_j <0.
\label{eq:flex}
\end{equation}
Due to the positiveness of the Lagrange multipliers, \eqr{eq:sustlearn_positivealpha}, it is easy to show that \eqr{eq:flex} attains the smallest values when $\sum_i [y_i=1] = \sum_i [y_i=-1]$, where $[\cdot]$ is the Iverson bracket. Having balanced training sets has also been experimentally certified to be beneficial in large scale supervised learning~\cite{good_practices}.
%
%


\subsection{Zero-shot priors}
In zero-shot learning the likelihood of an image $\mb{x}$ being classified as the unseen label $c$
is generally expressed as a linear combination of conditional probability distributions over a 
set of known concepts $\mathcal{K}$~\cite{akata13cvpr,MensinkCVPR2014,conse}:
\begin{equation}
p(c | \mb{x}) = \sum_{k \in \mathcal{K}} \beta_{ck}  \,  p(k | \mb{x}, \mb{w}_k),
\label{zeroshot}
\end{equation}
where $\mb{w}_k$ is the classifier for the $k$-th concept and the weights $\beta_{ck}$ indicate the relation between the known concepts and the new class. 


By using linear models, the zero-shot prediction can be written as 
\begin{equation}
 f^{zs}(\mb{x}) = \sum_{k \in \mathcal{K}} \beta_{ck} \, \mb{w}_k\cdot\mb{x}_i~,
\end{equation}
and exploited as an auxiliary source of knowledge while performing active learning.
Specifically we modify the prediction score of active learning to
\begin{equation}
f^{t}(\mb{x}) = \eta^{t}f^{zs}(\mb{x}) + \mb{w}^{t}\cdot\mb{x}.
\end{equation}
In this way, zero-shot learning is used as initialization at $t=0$ and supports the active learning process at each following step both in the training sample selection and in the test prediction.

\section{Query sampling Procedure} \label{sub:sustsampling}

\begin{figure}[t]
    \centering
    \includegraphics[width=.85\columnwidth]{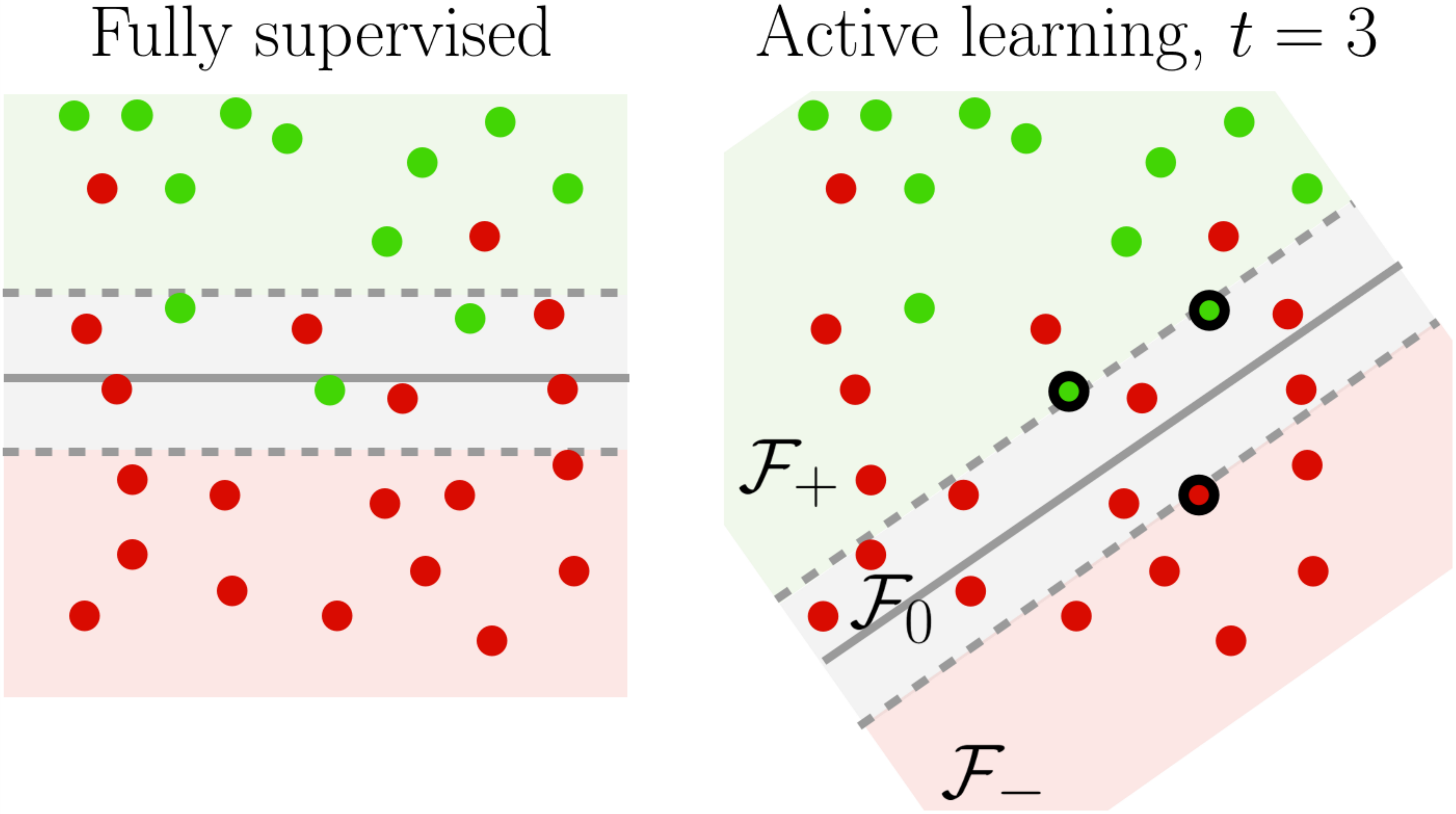}
    \caption{Sampling regions for SVM classifiers. Left: in a fully supervised SVM we expect that the (gray) margin-hyperplane zone $\mathcal{F}_0$  contains roughly as many positive as negatives, that are also the most confusing samples for the classifier. Right: as the active SVM classifier is uncertain in the beginning, sampling exclusively from $\mathcal{F}_0$ will likely result in negative labels during the early iterations, thus delaying the learning. We suggest sampling adaptively from the (green) positive outer margin zone $\mathcal{F}_+$ and $\mathcal{F}_0$ to maintain a good label distribution during active learning.}
    \label{fig:samplingfig}
\end{figure}

In the iterative process the current classifier divides the feature space into three zones:
the negative outer margin zone $\mathcal{F}_-$, the margin-hyperplane zone $\mathcal{F}_0$, and the positive outer margin
zone $\mathcal{F}_+$, see Fig. \ref{fig:samplingfig}. The samples will therefore be queried from one of the three zones. 

\subsection{Sampling from different feature space zones}
First, for balanced multiclass image collections as most often in computer vision, the prior class probabilities 
are roughly bounded: \ie $p(c) - p(c') < \delta$ for all categories $c, c'$ and an arbitrary small number $\delta$. 
As a result, by focusing on binary problems, we can safely expect that 
\begin{equation}
p(c) \ll p(\neg c),
\label{class-inequality}
\end{equation}
and thus most of the data will have negative labels.

Second, the performance of the defined active learning algorithm should be decent enough at the beginning due to the used zero-shot prior and it will progressively improve in time. 
We denote the random variable for a correct prediction, be it correctly \emph{positive} or correctly \emph{negative}, with $l$ and with $\neg l$ otherwise. 
Hence, we can expect the likelihood of a correct prediction to  be relatively higher than that of a wrong prediction
\begin{equation}
\frac{p(l]| \mb{w}^{t-1}, \mb{x})}{p(\neg l|\mb{w}^{t-1}, \mb{x})} > 1+\delta^{t-1}, 0 < \delta^{t-1} < \delta^{t}, \forall t\geq0.
\label{classifier-reliability}
\end{equation}
We consider the three zones and their influence of the sampling.

\textbf{Sampling from the negative outer margin zone $\mathcal{F}_-$.} 
The negative outer margin zone is defined by the region where $f^{t-1}(\mb{x}) < -1$. 
It is the largest zone,  according to \eqr{class-inequality}, and with a high probability of sampling real negatives $y_i=-1$, according to \eqr{classifier-reliability}.
Thus, even when sampling randomly in this region we are likely to repeatedly sample correctly classified negative examples, violating the label equality condition. Hence, we expect sampling from the negative outer margin zone to be suboptimal.

\textbf{Sampling from the hyperplane zone $\mathcal{F}_0$.} This region is defined by $-1 \leq f^{t-1}(\mb{x})  \leq +1$. It is the region where the classifier is maximally confused, and therefore is generally considered to be the best sampling zone~\cite{Schohn, Vijayanarasimhan2011}.
.
However, the samples close to the hyperplane have by definition low classification scores. Therefore, we expect that these data will have negligible effect, positive or negative, on the maximum conflict condition. Moreover, sampling from this region implicitly puts a lot of faith on the current hyperplane, an over-optimistic assumption. As a result, considering \eqr{class-inequality}, it is likely that many more negatives than positives will be sampled, which clashes with the label equality condition.

\textbf{Sampling from the positive outer margin zone $\mathcal{F}_+$.} This zone is defined by $f^{t-1}(\mb{x}) > 1$. 
In the early rounds our classifier is decent, yet not fully reliable. The likelihood of sampling a positive label will be only slightly higher than sampling a negative, but certainly higher than sampling a positive label from either $\mathcal{F}_-$ or $\mathcal{F}_0$.
If the new label is negative, $y_i=-1$, given that $f^{t-1}(\mb{x}_i) \gg 1$, and the signs are opposite, the maximum 
conflict condition is fully satisfied. If the label is positive, $y_i=+1$, the label equality condition gets closer to be satisfied over time with the abundance of negative samples.
Overall we conclude that sampling from the positive outer margin zone is beneficial for both the maximum conflict condition as well as the label equality condition. Hence we expect fastest learning in the first rounds when sampling from this zone.

\subsection{Maximum conflict-label equality sampling}
Sampling only from the $\mathcal{F}_+$ is beneficial at the early stages of active learning, when the reliability of the classifier can be low. In order to meet the maximum conflict and label equality conditions on later iterations, however, we propose a novel sampling strategy.

We consider the likelihoods for sampling positive and negative labels from $\mathcal{F}_+$ and $\mathcal{F}_0$, namely $p(l|\mathcal{F}_+, t)$, $p(\neg l|\mathcal{F}_+, t)$, $p(l|\mathcal{F}_0, t)$, $p(\neg l|\mathcal{F}_0, t)$. We compute these likelihoods at each iteration $t$, based on the previously sampled labels and the zones they were sampled from.

Assuming for clarity an annotation budget $B=1$, at iteration $t$ we have that
\begin{equation}
p(l|\mathcal{F}_+, t) \sim p(l|\mathcal{F}_+, t-1) + \frac{\sum_{r=1}^{t-1} [l_r=1]\cdot[\mb{x}^r \in \mathcal{F}_+^r]}{t-1},
\end{equation}
where $[\mb{x}^r \in \mathcal{F}_+^r]=1$ means that at iteration $r$ the sample $\mb{x}^r$ was sampled from the $\mathcal{F}_+$. 
The rest of the probabilities are computed in a similar manner. 
The term $p(l|\mathcal{F}_+, t-1)$ is included such that the likelihoods do not fluctuate violently, 
especially during the early iterations. Given that we have no information at the first round, for $t=1$ we set $p(l|\mathcal{F}_+, t-1)=0.5$ and $p(l|\mathcal{F}_0, t-1)=0.1$, although different 
initializations did not have any significant effect. At each iteration we normalize the probabilities to sum up to one, and we 
measure the label equality for the sampled data up to that point as
\begin{equation}
\rho^{t-1}=\frac{\sum_{r=1}^{t-1} [l_r=1]}{\sum_{r=1}^{t-1} [l_r=1]+[l_r=-1]}~.
\end{equation}
We then sample such that
\begin{equation}
\rho^{t-1} < \rho' \Rightarrow \mb{x}_t \sim \mathcal{F}_+^{t-1}~,
\end{equation}
namely we sample from $\mathcal{F}_+^{t-1}$ if we have too many negatives, otherwise we sample from $\mathcal{F}_0^{t-1}$. 
Selecting data from other regions, \eg negatives $\mathcal{F}_-^{t-1}$, did not seem to perform well in practice. Next, we present and discuss the empirical results of our adaptive active learning  approach.\\

\noindent\textbf{Empirical observations.}
The threshold $\rho'$ can be cross validated in a separate dataset, although a $\rho'=0.5$ seemed to work well in practice. Similar to MCMC Gibbs samplers~\cite{meyn1993markov} a burn-in period, where we sample only from $\mathcal{F}_+$, was shown to result in higher robustness. After enough samples have been queried, \eg about 150, the priors are not needed anymore, as we are in a discriminative setting, and thus they are dropped. Futhermore, we observe that in the first rounds most of the highest scoring samples are located in the edge between $\mathcal{F}_+$ and $\mathcal{F}_0$. This is because the training set is still small and also, because of the well-known shift for positive and negative sample scores, when moving from the train to the test set. Still the MCLE sampling strategy remains consistent, as we query samples after ranking, namely querying from $\mathcal{F}_+$ equals to querying the most confidently positive sample and querying from $\mathcal{F}_0$ equals to querying the most confused new sample.

\section{Experiments}


\begin{figure}[t]
    \centering
    \includegraphics[width=0.95\linewidth]{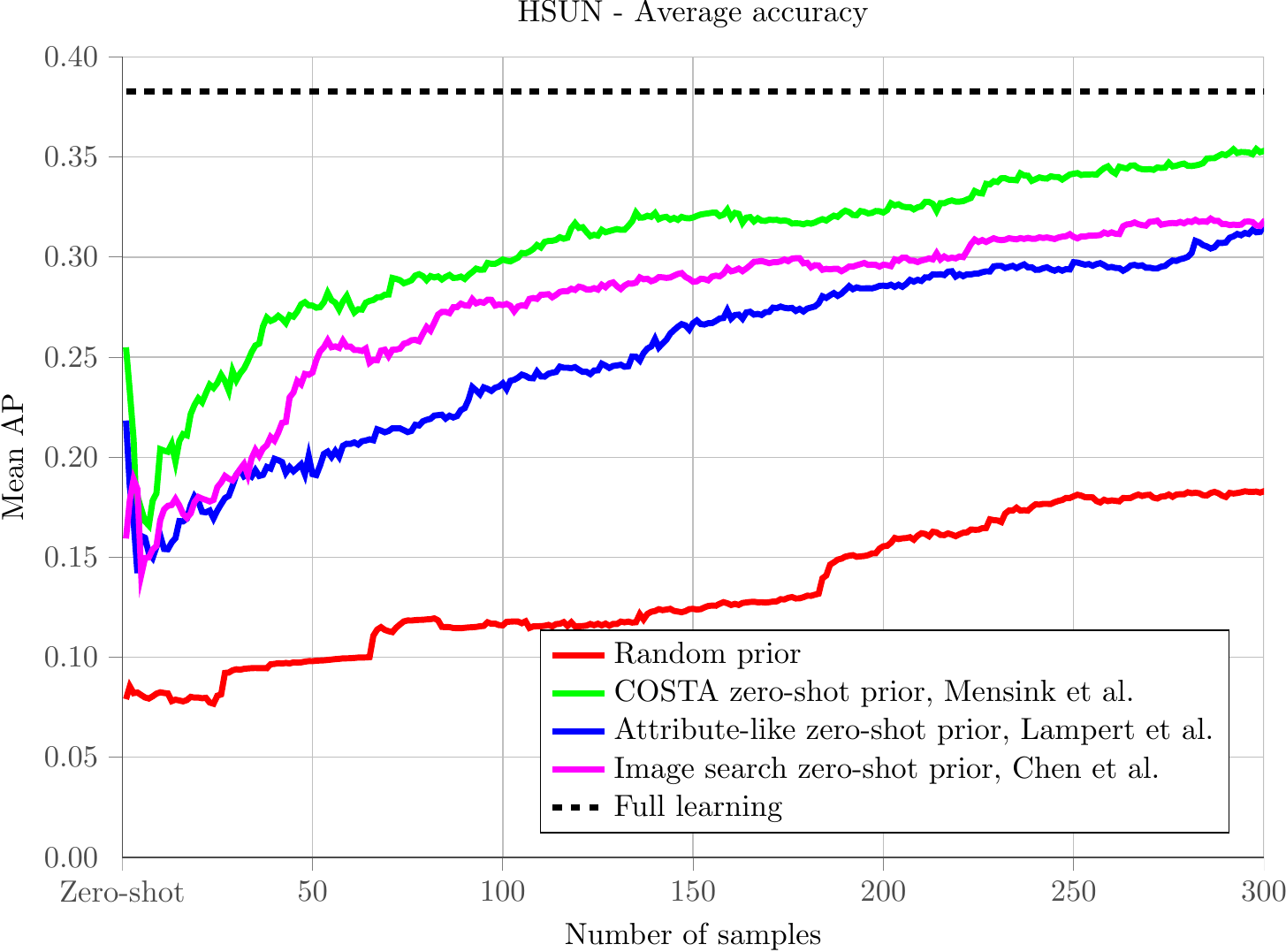}\\
    \includegraphics[width=0.95\linewidth]{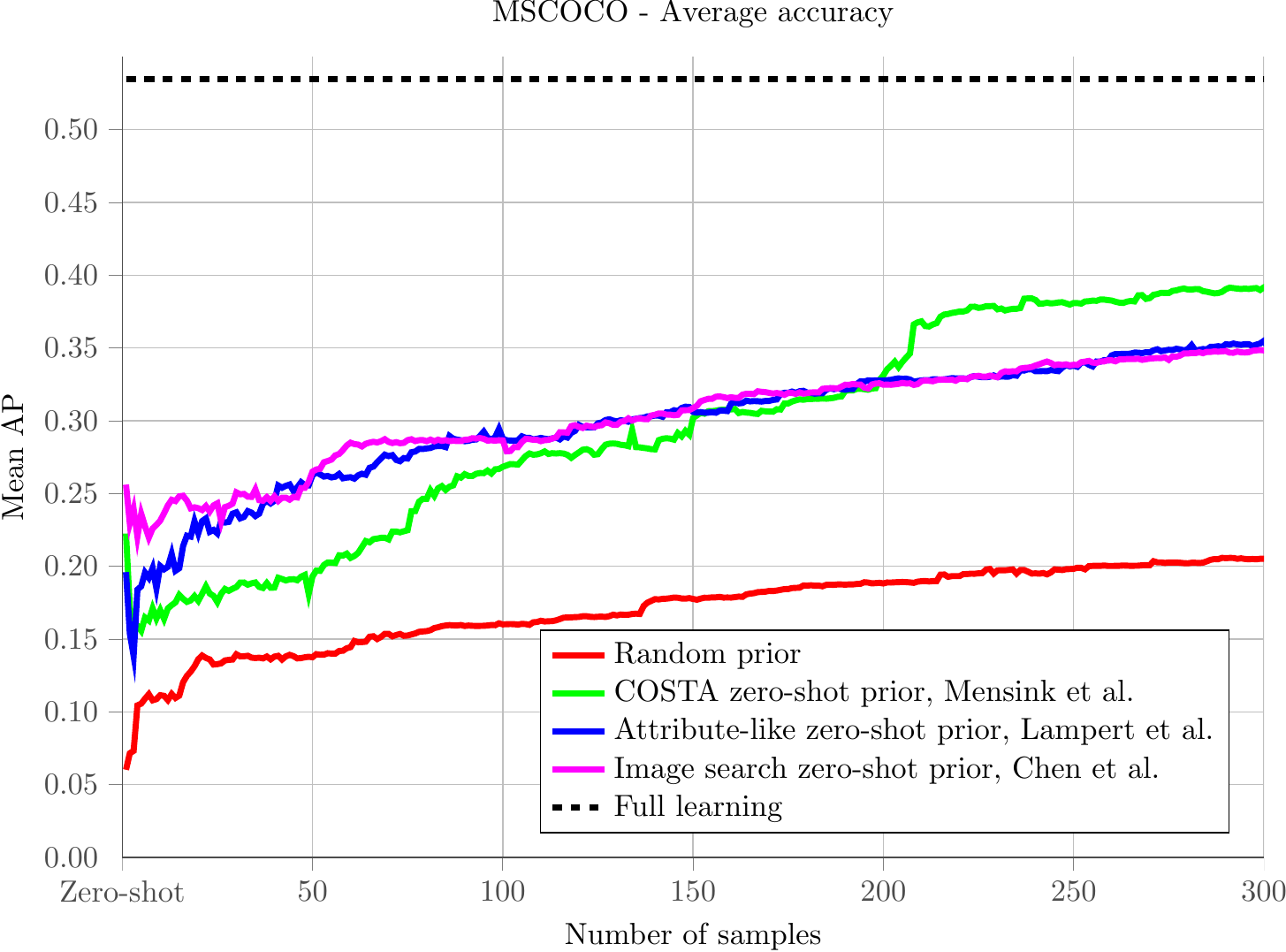}
    \caption{Various sources of prior information for active learning on HSUN (\emph{top}) and MSCOCO. The reason for the lower accuracy in MSCOCO is that we use only 300 out of the total $80,000$ training images (for HSUN there are $4,000$ in total).
    Active learning with zero-shot priors considerably outperforms a non-informative random prior, the default in the active learning literature.}
    \label{fig:priors}
    \figurecorrection
\end{figure}

\subsection {Experimental setup}
\label{sec:EXPsetup} 
We run \footnote{We implemented the MCLE zero-shot active learning in MATLAB. The code is available at http://www.egavves.com/list-of-publications.} our experiments on two challenging image datasets. \textbf{Hierarchical SUN (HSUN) dataset~\cite{ChoiCVPR2010}.}  This is a generic, multi-class and multi-label image classification dataset, covering both object as well as scene categories. There are in total 107 classes and 8,634 images, split into a training and a test set composed of 4,367 and 4,317 images respectively. \textbf{Microsoft COCO (MSCOCO) dataset~\cite{LinECCV2014}.} MSCOCO is a multi-class and multi-label dataset containing 80 object categories. The MSCOCO dataset contains 123,287 images in total, split into a training and a test set composed of $82,783$ and $40,504$ images respectively. Each image is annotated with respect to the presence or absence of a particular object. 

Following the common procedure in the zero-shot recognition literature~\cite{akata13cvpr, MensinkCVPR2014}, we divide 
the classes randomly into two sets: 75\% of known and 25\% of unknown classes. Also, all training and active 
sample querying is performed strictly on the training set. The evaluations, as well as the reported results
(Mean Average Precision: Mean AP), are computed using the completely independent test sets.
For both datasets and all experiments we rely on deep learning features~\cite{KrizhevskySH12}, trained on 
the separate \textit{ImageNet} dataset~\cite{imagenet_cvpr09}. We fix the SVM learning parameter $C$
to $1$ and we run our method over 300 iterations (\ie the maximum number of queried training samples is 300).

%

\begin{figure}[t!]
    \centering
    \includegraphics[width=0.95\linewidth]{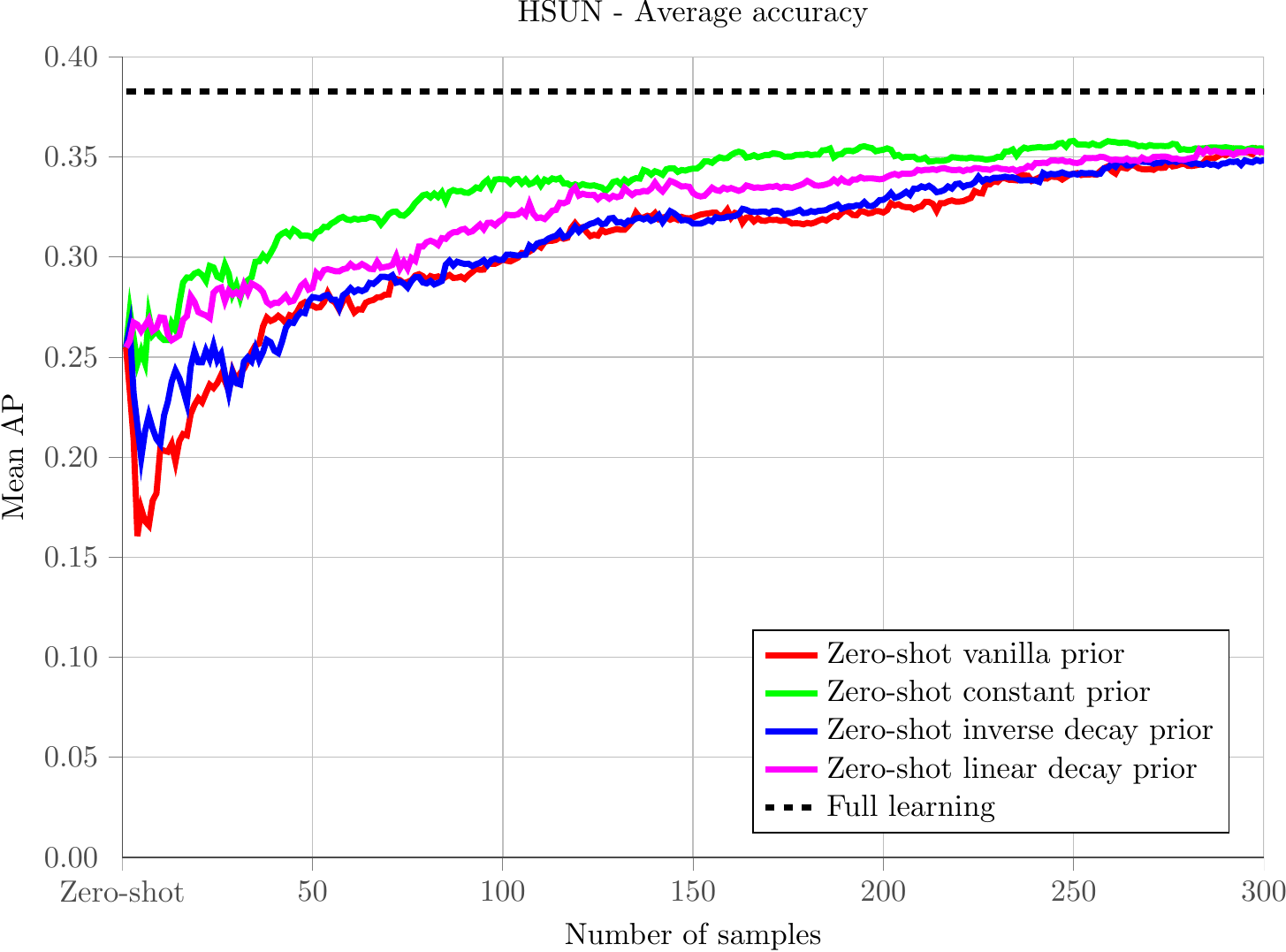}
    \caption{Different ways to update the zero-shot prior for active learning. A constant prior works best. Moreover, zero-shot learning acts as a good initialization and accelerates the active learning.}
    \label{fig:prior-forms}
    \figurecorrection
\end{figure}

\subsection{Zero-shot priors for active learning} \label{sec:EXPzeroshot}
We first establish the value of using a zero-shot model as an auxiliary source for active learning. To that end, we compare the effect of the zero-shot warm initialization against a basic random initialization which is a standard choice in active learning. Regarding sampling we opt for simplicity and sample from $\mathcal{F}_+$.
We compare three zero-shot models: 
\textbf{COSTA}~\cite{MensinkCVPR2014} using co-occurrences to express the class-to-class relation;
\textbf{Attribute-like} models using binarized class-to-class relations mined from other sources~\cite{lampert13pami,rohrbach10cvpr}; and \textbf{Image search} model based using the first 12 images returned by Google image search as positive examples, similar to~\cite{Chen_2013_ICCV}.
These zero-shot models are used as auxiliary model $f^{a}(\cdot)$ only in the first step of the active learning process to sample the first example and initiate SVM learning, \ie the classifier is $f(\mb{x})^t = \eta^t f^{a}(\mb{x}) + \mb{w}^t\cdot\mb{x}$ with $\eta^0=1$ and $\eta^t=0$ for $t>0$.

The obtained results are presented in Fig.~\ref{fig:priors}, including the fully supervised results as upper bound.
We observe that using a zero-shot warm initialization always improves the learning considerably, both at $t=0$ and at $t=300$. For the HSUN dataset the COSTA prior is best from start to end. For the object-oriented MSCOCO dataset the image search zero-shot prior is better in the first iterations, whereas COSTA is better afterwards.

In a second experiment, we evaluate different ways to exploit the zero-shot prior during the active learning process.
We compare 4 strategies: \textbf{vanilla prior} where we use the prior only at $t=0$ as above, \textbf{constant prior} where $\eta^t=1~, \forall t$, \textbf{inverse decay prior} where $\eta^t=1/(t+1)$, and finally \textbf{linear decay prior} where $\tfrac{t_0}{t+t_0} f^{a}(\mb{x}) + \tfrac{t}{t+t_0}\mb{w}^t\cdot\mb{x}$ with a relatively high $t_0=20$.
%
The results are presented in Fig.~\ref{fig:prior-forms}. 
The constant prior appears to be the fastest learner, proving that zero-shot learning works not only as a good initialization but can also consistently accelerate the active learning. As expected, all the considered variants converge when a larger number of training samples is available.
On the basis of these results, we use in the remaining experiments the constant prior, using COSTA source models for HSUN and the Image search source models for MSCOCO.
\begin{figure}[t!]
    \centering
    \includegraphics[width=0.95\linewidth]{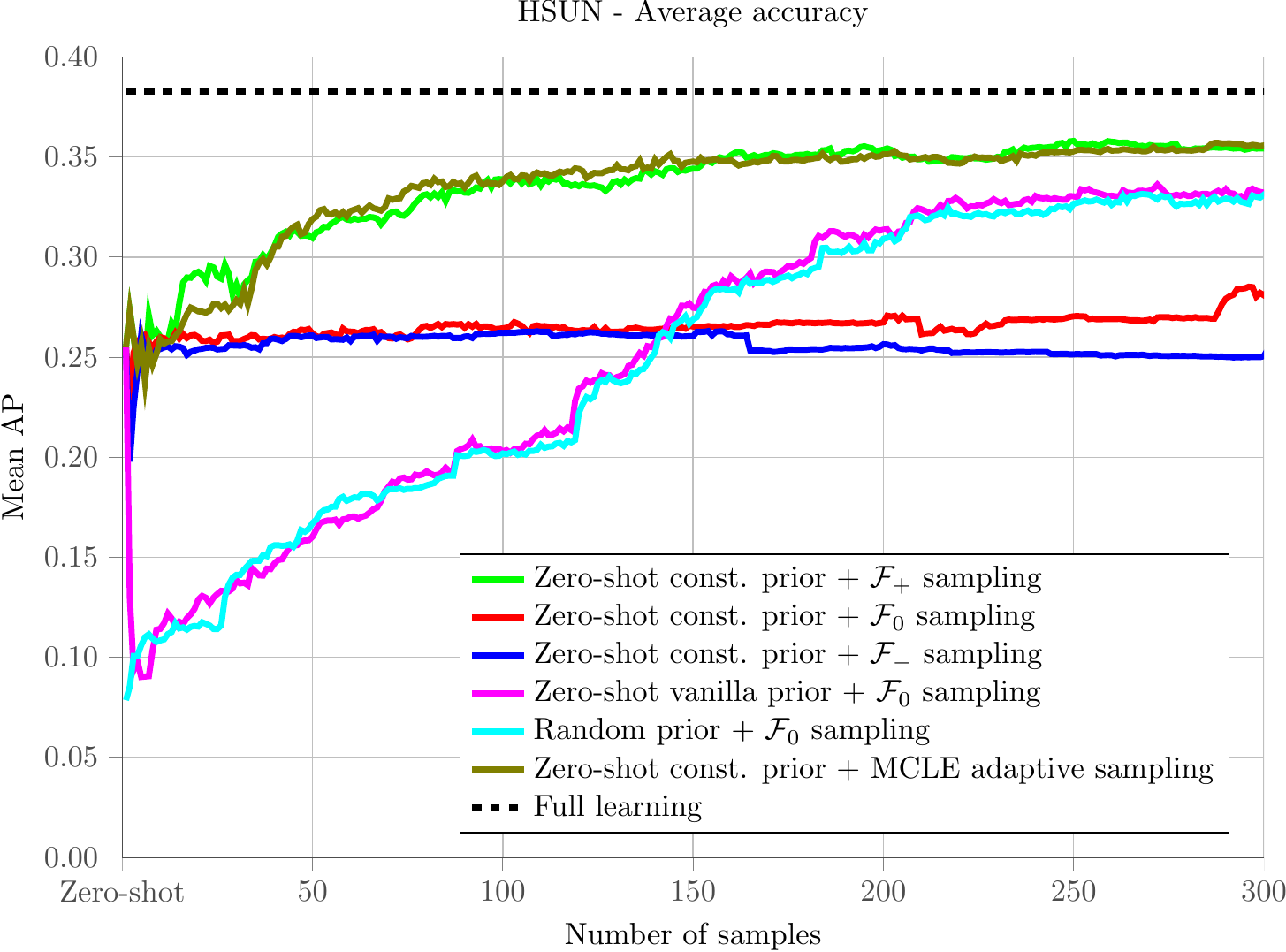}\\
    \includegraphics[width=0.95\linewidth]{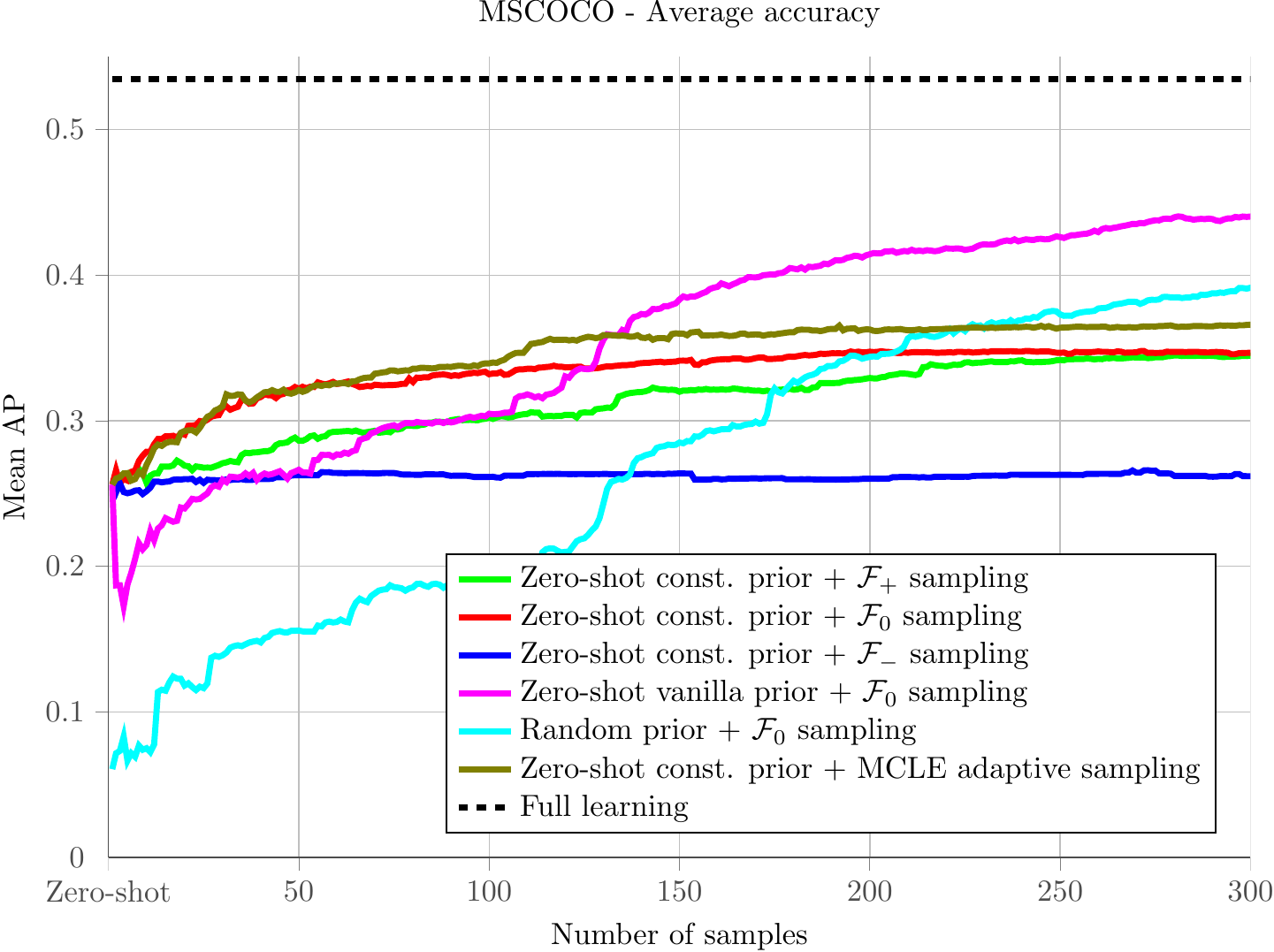}
    \caption{Different active sampling strategies for HSUN (\emph{top}) and MSCOCO. The adaptive MCLE sampling works as good as $\mathcal{F}_+$ for HSUN and considerably better for MSCOCO.}
    \label{fig:active-sampling}
    \figurecorrection
\end{figure}

\subsection{Maximum conflict-label equality} \label{sec:EXPmin-slack}
In this experiment we evaluate different sampling strategies as detailed in Sec.~\ref{sub:sustsampling}. 
Note that sampling from $\mathcal{F}_0$ is equivalent to selecting the data points on which the current
classifier presents the maximal uncertainty, as repeatedly proposed in the literature~\cite{Tong2002, Schohn, Vijayanarasimhan2011}. 
We present results for the HSUN and the MSCOCO datasets in Fig.~\ref{fig:active-sampling}.

For the HSUN dataset the zero-shot priors might be weak (COSTA co-occurrence classifiers),
while for MSCOCO the priors are more reliable (image search supervised classifiers). Moreover,
in HSUN the images are harder to classify than in MSCOCO as can be appreciated by comparing the accuracy
of the fully supervised case. These observations explain the different behaviours visible in the
two plots. For HSUN, $\mathcal{F}_+$ contains both positive and negative samples with a roughly 
balanced distribition, and MCLE mainly selects samples from this region. For MSCOCO, $\mathcal{F}_+$ contains mainly positive
samples and, to have a balanced set, MCLE chooses samples also from $\mathcal{F}_0$. The results of the vanilla prior
baseline confirms that it is useful to sample from $\mathcal{F}_0$: an analysis on the 
accuracies of the individual class experiments reveales that the high increase in performance after 125 iterations
coincides with the optimal label equality.

Overall we can state that our method automatically adapts to the different conditions providing always higher 
or equal results than selecting only from $\mathcal{F}_+$ or $\mathcal{F}_0$. We conclude that MCLE sampling 
is a reliable strategy when using zero-shot priors for active learning.

\begin{table*}[t!]
  \centering
  \setlength{\tabcolsep}{6pt}
 \scalebox{0.8}{
  \begin{tabular}{l c c c c c c c c c c c c c c c}
  \toprule
  & \multicolumn{7}{c}{\textit{HSUN (All samples: 0.383 mAP)}} & & \multicolumn{7}{c}{\textit{Small COCO (All samples: 0.460 mAP)}} \\
  \cmidrule{2-8} \cmidrule{10-16} 
  No. of queries & 0 & 50 & 100 & 150 & 200 & 250 & 300 & & 0 & 50 & 100 & 150 & 200 & 250 & 300 \\
  \midrule
  MCLE (Dataset prior) & 0.255 & 0.315 & 0.337 & 0.348 & 0.346 & 0.355 & 0.361 & & 0.250 & 0.350 & 0.381 & 0.383 & 0.444 & 0.448 & 0.460 \\
  MCLE (External prior) & 0.270 & 0.289 & 0.327 & 0.341 & 0.336 & 0.348 & 0.358 & & 0.197 & 0.293 & 0.391 & 0.427 & 0.436 & 0.442 & 0.457 \\
  \midrule
  BBAL~\cite{VijayanarasimhanJG10} & 0.158 & 0.241 & 0.276 & 0.309 & 0.322 & 0.328 & 0.325 & & 0.168 & 0.283 & 0.335 & 0.364 & 0.380 & 0.395 & 0.408 \\
  Hiearchical Sampling~\cite{Dasgupta} & 0.089 & 0.156 & 0.199 & 0.221 & 0.234 & 0.230 & 0.246 & & 0.076 & 0.182 & 0.250 & 0.287 & 0.309 & 0.331 & 0.365 \\
  GP Mean~\cite{Kapoor} & 0.154 & 0.282 & 0.319 & 0.340 & 0.350 & 0.361 & 0.365 & & 0.186 & 0.344 & 0.394 & 0.412 & 0.421 & 0.431 & 0.438 \\
  GP Variance~\cite{Kapoor} & 0.154 & 0.201 & 0.206 & 0.216 & 0.226 & 0.240 & 0.244 & & 0.186 & 0.233 & 0.263 & 0.284 & 0.291 & 0.309 & 0.326 \\
  GP Impact Bayes~\cite{Freytag14} & 0.154 & 0.251 & 0.286 & 0.305 & 0.316 & 0.327 & 0.345 & & 0.186 & 0.298 & 0.346 & 0.393 & 0.417 & 0.430 & 0.436 \\
  GP EMOC Bayes~\cite{Freytag14} & 0.154 & 0.277 & 0.310 & 0.320 & 0.328 & 0.332 & 0.337 & & 0.186 & 0.336 & 0.375 & 0.388 & 0.397 & 0.399 & 0.405 \\
  \bottomrule
  \end{tabular}
  }
  \caption{Comparison with state-of-the-art active learning methods. For HSUN the external prior comes from COSTA learned on MSCOCO, whereas for MSCOCO the external COSTA prior is learned on HSUN. For both datasets, the MCLE sampling outperforms all baselines almost always, especially in the early rounds. In fact for Small MSCOCO MCLE reaches the full mAP within 300 samples, even with external priors. Hence, \emph{we effectively reuse past datasets}.} \label{tab:compare_al}
\vspace{5mm}
\end{table*}


\subsection{State-of-the-art comparisons} \label{sec:EXPsota}

\begin{table}[t!]
  \centering
  \setlength{\tabcolsep}{6pt}
  \scalebox{0.65}{
  
  \begin{tabular}{l l c c c c c c c c}
  \toprule
    & Class  & 0 & 50 & 100 & 150 & 200 & 250 & 300 \\
  \midrule
  MCLE & \multirow{2}{*}{\textit{Bicycle}} & 0.134 & 0.244 & 0.259 & 0.258 & 0.279 & 0.286 & 0.296 \\
  BBAL~\cite{VijayanarasimhanJG10} &  & 0.046 & 0.173 & 0.190 & 0.250 & 0.268 & 0.272 & 0.257 \\
  \midrule
  MCLE & \multirow{2}{*}{\textit{Stop sign}} & 0.048 & 0.307 & 0.313 & 0.316 & 0.369 & 0.369 & 0.375 \\
  BBAL~\cite{VijayanarasimhanJG10} & & 0.021 & 0.031 & 0.032 & 0.033 & 0.150 & 0.320 & 0.312 \\
  \midrule
  MCLE & \multirow{2}{*}{\textit{Cow}} & 0.019 & 0.046 & 0.127 & 0.269 & 0.275 & 0.283 & 0.352 \\
  BBAL~\cite{VijayanarasimhanJG10} &  & 0.166 & 0.206 & 0.204 & 0.258 & 0.281 & 0.299 & 0.300 \\
  \midrule
  MCLE & \multirow{2}{*}{\textit{Refrigerator}} & 0.365 & 0.379 & 0.419 & 0.453 & 0.474 & 0.452 & 0.479 \\
  BBAL~\cite{VijayanarasimhanJG10} &  & 0.319 & 0.291 & 0.407 & 0.341 & 0.375 & 0.381 & 0.384 \\
  \bottomrule
  \end{tabular}
  }
  \caption{Accuracies for specific categories on Small COCO. For MCLE we use external prior from HSUN COSTA. We compare with BBAL~\cite{VijayanarasimhanJG10} the best SVM based active learning method from Tab.~\ref{tab:compare_al}. We observe that the MCLE sampling learns faster and better even when the zero-shot priors are not good enough. Note that~\cite{VijayanarasimhanJG10} assumes there is at least one positive sample available at the beginning, a strong assumption for practical use.}
\label{tab:soa-cat}
\vspace{-5mm}
\end{table}


\begin{figure}[t!]
    \centering
    \includegraphics[width=0.95\linewidth]{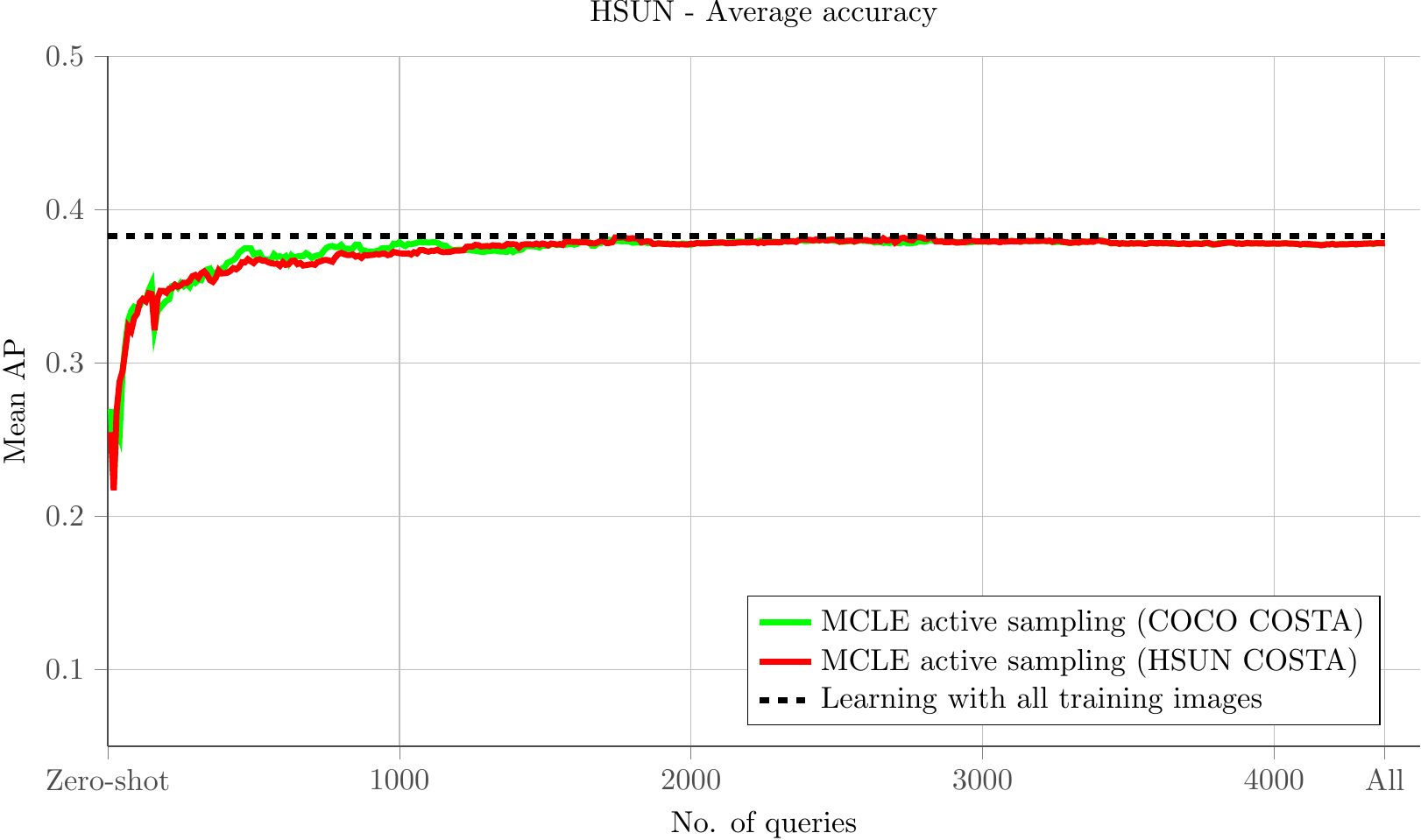}\\
    \includegraphics[width=0.95\linewidth]{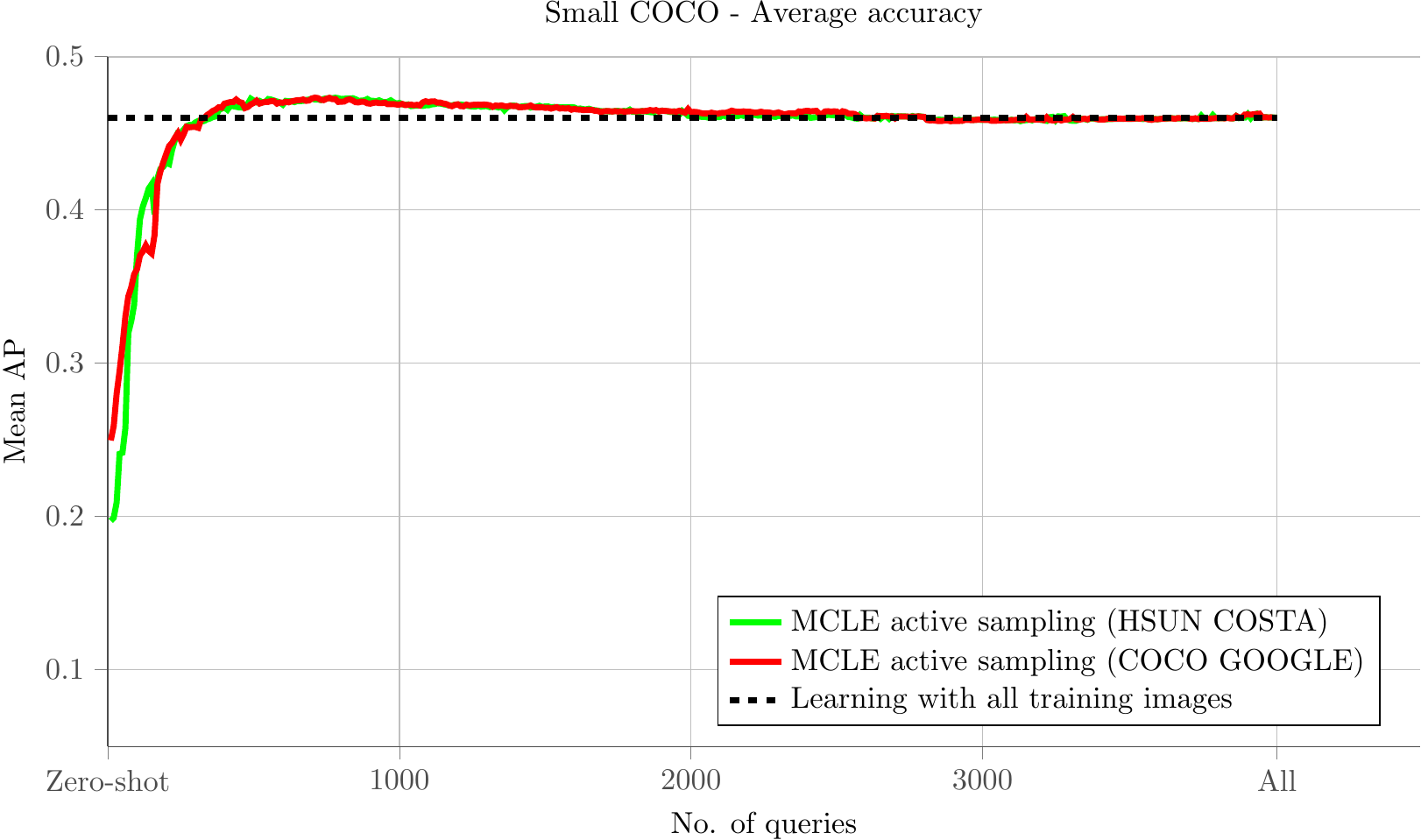}
    \caption{Active learning progress with MCLE when all training samples are used. MCLE sampling reaches the full accuracy rather early in the process. With only 300 samples, which is about 95\% of the labels MCLE obtains 95-100\% of the mAP for both datasets.}
\label{tab:soa}
    \label{fig:mcle-full}
    \figurecorrection
\end{figure}

Next, we present comparisons to state-of-the-art active learning methods. For computational reasons we perform all experiments on a random subset of MSCOCO containing 4,000 training and 4,000 test images, which we call `Small MSCOCO'. We consider the  \emph{far-sighted} active learning from~\cite{VijayanarasimhanJG10}, the hierarchical sampling approach~\cite{Dasgupta}, as well as the Gaussian Process based active learning methods from~\cite{Freytag14} and~\cite{Kapoor}. For these baselines we use the publicly available code and the recommended settings.

\begin{figure*}[t!]
    \centering
    \includegraphics[width=1.0\linewidth]{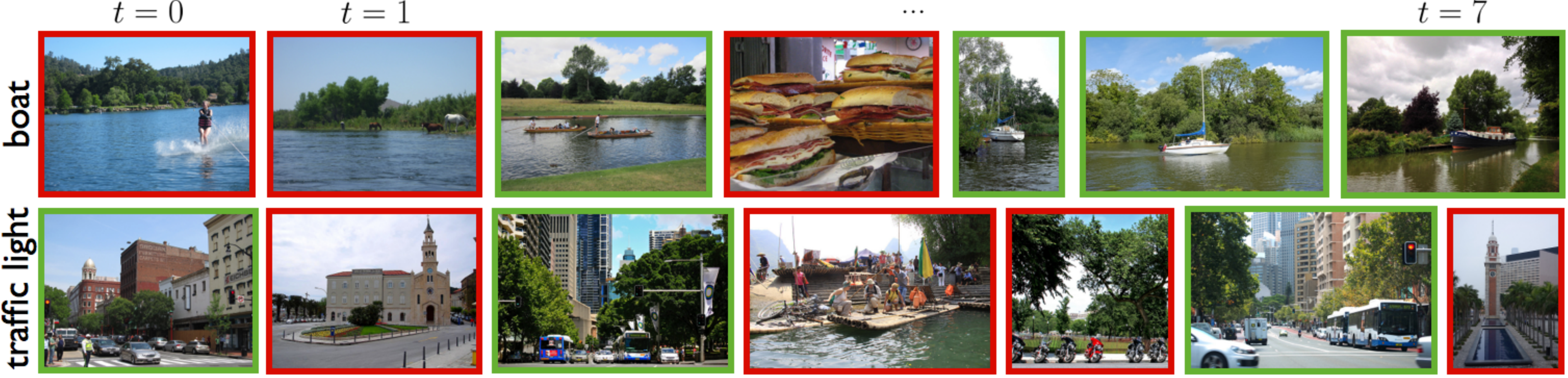}
    \caption{Images sampled during the first seven iterations of active learning for \emph{``boat''} and \emph{``traffic light''} on MSCOCO with HSUN zero-shot priors. Positive samples are denoted with green, negative ones with red. Observe the balanced label distribution, while the visual similarity indeed reveals the maximum conflict during the selection of the next sample.}
    \label{fig:examples}
    \vspace{5mm}
\end{figure*}

For MCLE sampling we use two types of priors: one from the training set of the same dataset which can be over-optimistic, and a second one from external data.
For HSUN we use the COSTA priors computed on MSCOCO, while for MSCOCO we use the image search classifiers. 
To build the label-to-label co-occurrence matrix for the zero-shot COSTA priors we 
compute the Flickr tag statistics as in~\cite{MensinkCVPR2014}. For sake of compatible comparisons, we perform the experiment on the same unknown categories (and not the full dataset) as in the previous experiments. 


All baselines require one positive and one negative sample 
at time $t = 0$, while they are not needed for MCLE
sampling with zero-shot priors. This provides a significant
advantage for the baselines: selecting a positive data out of
an unbalanced distribution with a large number of negatives
is challenging and surely a strong assumption. Still, we believe that useful conclusions can
be drawn. We present results in Tab.~\ref{tab:soa}.


MCLE sampling with zero-shot priors outperforms all other methods, especially in the early rounds where classifiers are most uncertain.
For Small MSCOCO MCLE sampling allows for obtaining even the full mAP with only 300 samples.
Furthermore, MCLE sampling with external priors is as good as when using priors from the own dataset.

In Tab.~\ref{tab:soa-cat} we present results for separate categories in Small MSCOCO, using the MCLE with external prior and the best SVM-based active learning method from the results above.  MCLE sampling learns faster and better even for poor zero-shot priors.
Furthermore, with active learning on HSUN using MSCOCO zero-shot priors, 10 out of the 26 unknown labels are common between the two datasets (for the reverse 4 labels are common). Hence, results remain good even in the simultaneous presence of known and unknown labels. To rule out negative effects of severe label imbalance, we repeat the experiments for BBAL~\cite{VijayanarasimhanJG10} tuning the misclassification penalty $C$ per class. We observed no benefit. Further, the hierarchical sampling~\cite{Dasgupta} intrinsically deals with label imbalance, still giving inferior results. We conclude that low accuracy is due to the query samples chosen rather than label imbalance.
Finally, we plot in Fig.~\ref{fig:mcle-full} the full evolution of the active learning until all training images are used. Full accuracy is obtained quite fast, thus allowing for a considerable saving in annotation requirements: with 300 samples, namely about 5-10\% of the labels, we obtain about 95-100\% of the full mAP. 

Based on the \emph{maximum conflict-label equality} conditions we can exploit past datasets and existing human annotations for the active learning of future tasks on unknown categories. We conclude that zero-shot priors and MCLE sampling allows for a faster, more robust and evidently a more economical active learning. To the best of our knowledge, \emph{we are the first to show how to exploit existing but unrelated annotations for active learning of novel image categories.}

\section{Conclusion}
In this work we attempt a first answer to the question of how to reuse past datasets for 
faster and more accurate learning of new and seemingly unrelated future tasks. We start from 
zero-shot classifiers and re-purpose them as priors to warm up active learning. 
Focusing on the dual formulation of SVM, we reveal two conditions for optimal sampling, 
identifying the most important samples to be annotated next. We then propose an 
effective active learning approach that learn the best possible target classification model 
with minimum human labeling effort. 

Up to our knowledge no previous work combined zero-shot with active learning.
As demonstrated by the MCLE optimal sampling conditions, this setting is different from standard
active learning without auxiliary knowledge, in that the positive outer margin zone $\mathcal{F}_+$
gains importance with respect to the traditional hyperplane zone $\mathcal{F}_0$.

Extensive experiments on two challenging datasets show the value of our approach compared to the state-of-the-art, and outline the potential of reusing past datasets with minimal effort for future recognition tasks.

\vspace{2mm}
{
\small
\textbf{Acknowledgments}
This research is supported by the iMinds strategic project HiViz, the STW STORY project and the Dutch national program COMMIT.
}

\clearpage
{\small
\bibliographystyle{ieee}
\bibliography{sustainable_learning}
}

\end{document}